\documentclass[runningheads]{llncs}
\usepackage{graphicx}
\usepackage{subfigure}
\usepackage{makecell}
\usepackage{cite}

\begin{document}

\title{CubemapSLAM: A Piecewise-Pinhole Monocular Fisheye SLAM System\thanks{This work is partially supported by the National Natural Science Foundation (61872200), the Natural Science Foundation of Tianjin (17JCQNJC00300) and the National Key Research and Development Program of China (2016YFC0400709).}} 
\titlerunning{CubemapSLAM} 


\author{Yahui Wang\inst{1}
\and
Shaojun Cai\inst{2}
\and
Shi-Jie Li\inst{1}
\and
Yun Liu\inst{1}
\and
Yangyan Guo\inst{3}
\and
Tao Li\inst{1}
\and
Ming-Ming Cheng\inst{1}
}
%

\authorrunning{Y. Wang et al.} 


\institute{The College of Computer and Control Engineering, Nankai University, China 
\and
UISEE Technology (Beijing) Co., Ltd. \\
\and
University of Chinese Academy of Sciences, China\\
}

\maketitle

\begin{abstract}
We present a real-time feature-based SLAM (Simultaneous Localization and
Mapping) system for fisheye cameras featured by a large field-of-view (FoV).
Large FoV cameras are beneficial for large-scale outdoor SLAM applications, because they increase visual overlap between consecutive frames and capture more pixels belonging to the static parts of the environment. However, current feature-based SLAM systems such as PTAM and ORB-SLAM limit their camera model to pinhole only. 
To compensate for the vacancy, we propose a novel SLAM system with the cubemap model that utilizes the full FoV without introducing distortion from the fisheye lens, which greatly benefits the feature matching pipeline. 
In the initialization and point triangulation stages, we adopt a unified vector-based representation to efficiently handle matches across multiple faces, and based on this representation we propose and analyze a novel inlier checking metric. In the optimization stage, we design and test a novel multi-pinhole reprojection error metric that outperforms other metrics by a large margin. We evaluate our system comprehensively on a public dataset as well as a self-collected dataset that contains real-world challenging sequences. 
The results suggest that our system is more robust and accurate than other feature-based fisheye SLAM approaches. The CubemapSLAM system has been released into the public domain.

\end{abstract}
\section{Introduction}
SLAM techniques have been widely applied in the robotics and automation industry. Specifically, Visual SLAM (\textbf{VSLAM}) is gaining increasing popularity, 
because cameras are much cheaper than other alternatives such as differential GPS (D-GPS) and LIDAR.
However, traditional VSLAM systems suffer from problems such as occlusions, moving objects and drastic turns due to the limited FoV of perspective cameras. 
In contrast, large FoV cameras significantly increase the visual overlap between consecutive frames. In addition, large FoV cameras capture more information from the environment, therefore making the SLAM system less likely to fail.

However, there are still many challenges in SLAM with large FoV cameras. The first challenge is that most of the widely-used feature descriptors are designed for low-distortion images. Some systems \cite{tardif2008monocular,scaramuzza2008appearance,rituerto2010visual} choose more robust features such as SIFT\cite{lowe2004distinctive} or design new features suited for highly distorted images \cite{zhao2015sphorb,arican2010omnisift}, but they are too time-consuming to satisfy the real-time demands of many applications.
Others \cite{furgale2013toward,lee2013motion,heng2013camodocal} try to remove distortion effect by directly rectifying fisheye images into pinhole images, but the remaining FoV is much smaller after rectification.
Multicol-SLAM\cite{urban2016multicol} adapts ORB-SLAM\cite{mur2015orb} to operate on the raw distorted images, but the open-source version fails to achieve satisfying results.

In this paper, we redesign the pipeline of ORB-SLAM to fit the piecewise linear camera model that utilizes full FoV without introducing distortion. 
We thus propose an efficient and compact feature-based SLAM system dedicated to large FoV cameras. 
Our system achieves better performance than directly rectifying the fisheye image into a pinhole image and the other existing feature-based fisheye SLAM system \cite{urban2016multicol}. Despite the limited angular resolution of a fisheye camera, we achieve comparable accuracy to ORB-SLAM with a pinhole camera while performing much more robustly. Specifically, our work has the following contributions:
\begin{enumerate}
	\item We propose the first cubemap solution for feature-based fisheye SLAM. 
	The piecewise-pinhole nature of the cubemap model is especially desirable for feature descriptors, and there is no need to retrain Bag-of-Words (\textbf{BoW}) \cite{galvez2012bags} vocabulary for fisheye images. 
	
	\item In the initialization and point triangulation stages, we adopt a unified vector-based representation which efficiently handles the matches across multiple faces. Based on this representation, we propose a novel and systematic inlier checking metric for RANSAC with essential matrix constraint, and we provide a rigorous analysis of the correctness of this metric. 
	
	\item In the optimization stage, we carry out thorough comparisons of different error metrics, and we propose a novel reprojection error metric for the cubemap model that outperforms other metrics.
	\item We present an extensive evaluation on public datasets, and a self-collected one containing typical outdoor driving scenarios. We also discover that by carefully choosing the camera mounting position, the problem of a low angular resolution in outdoor scenes mentioned in \cite{zhang2016benefit} can be greatly reduced.
\end{enumerate}

\subsection{Related Work}
The VSLAM techniques have been widely used in various applications such as self-driving cars \cite{ros2012visual, ziegler2014making,linegar2015work}. 
However, limited FoV of pinhole camera may cause the localization system to fail when there is little overlap between consecutive frames. Consequently, large FoV cameras are gaining attention. For instance, the V-Charge project \cite{lee2013motion,furgale2013toward} builds a car surrounded with 4 synchronously triggered fisheye cameras modeled as a generalized camera \cite{pless2003using}. 
In recent years, many works have discussed the methods to exploit large FoV cameras. 
\cite{kangni2007orientation} transforms the panoramas from the PointGrey LadyBug camera into cubic panoramas, 
but they aim to estimate poses of input cubic panoramas rather than build real-time SLAM system. 
A piecewise-pinhole model is presented in \cite{ventura2012wide}, but in the work the map needs to be built offline, and the local and global bundle adjustment as well as the loop closing based on the proposed model are not performed.
A number of semi-direct or direct SLAM systems based on fisheye models have been proposed recently \cite{heng2016semi,liu2017direct}. In their works an adapted GPU-based plane-sweep stereo algorithm is used to find matching patches between stereo image from the raw fisheye images. Omni-LSD \cite{caruso2015large} also proposes a similar pinhole array model as part of the extension to origin LSD-SLAM \cite{engel2014lsd}. While direct method is shown to be robust in scale-diverse environment, 
its performance in large outdoor environments is still unknown. To our knowledge, Multicol-SLAM\cite{urban2016multicol} is the only existing feature-based fisheye SLAM, but it tries to extract features directly on highly distorted images, which may lead to false matches.
Further comparison with MultiCol-SLAM will be presented in the experiment section. 

In this work, we propose an efficient and practical cubemap SLAM solution aimed at large-scale outdoor applications. In the following sections, we will first introduce the theoretical adaptions we have made in order to maximally utilize the power of a cubemap model, and then we will demonstrate the advantage of our system in extensive large real-world experiments.

\begin{figure}
	\centering
	\includegraphics[width=0.8\textwidth]{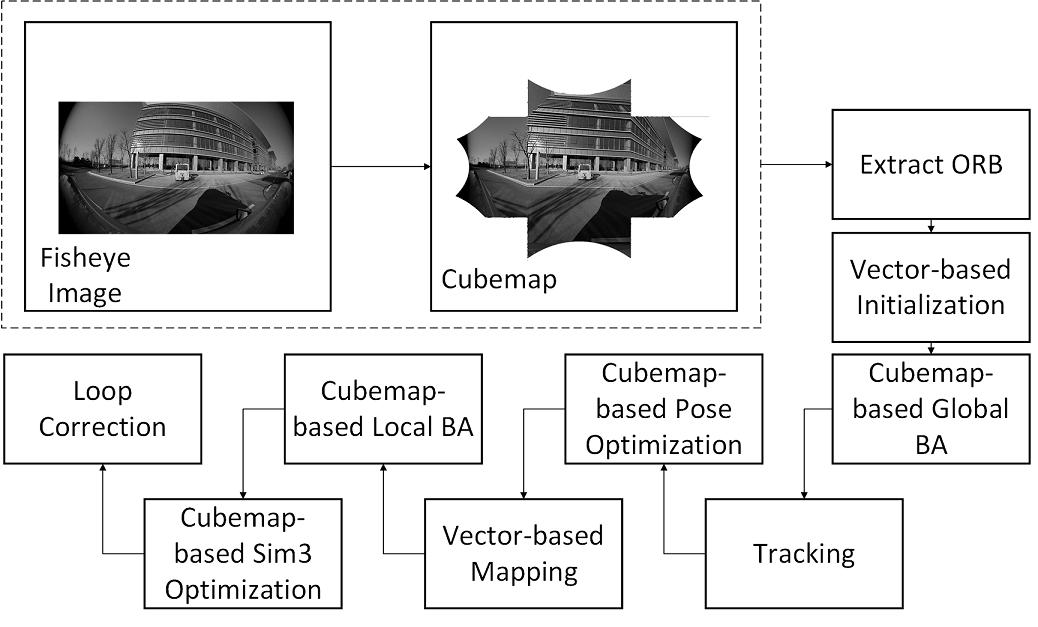}
	\caption{System overview}
	\label{fig_system_overview}
\end{figure}
\section{Algorithm}
\label{sec_algorithms}
In this section, we describe the pipeline of the proposed method. 
As shown in Fig. \ref{fig_system_overview}, we acquire cubemap image by calibrating the fisheye camera and mapping fisheye images onto a cube. 
A vector-based RANSAC is used to solve the essential matrix to recover camera motion and build the initial map. 
A cubemap-based global bundle adjustment is used to refine camera poses and the initial map. After initialization, the tracking thread estimates camera poses by tracking the local map and refines the poses with a cubemap-based pose optimization algorithm. 
When the tracking thread decides to insert current frame into the map as a keyframe, a vector-based triangulation algorithm is used to create new map points, and the frame is converted into BoW vectors for loop detection. When a loop is detected, a $Sim_3$ transformation for loop closing is computed by an adapted $Sim_3$ optimization algorithm and a loop correction is performed.

\begin{figure}
	\centering
	\subfigure[]{
		\label{fig_bearing_vectors}
		\includegraphics[scale=0.32]{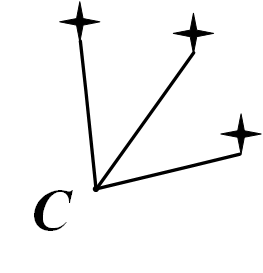}}
	\subfigure[]{
		\label{fig_mapping_to_plane}
		\includegraphics[scale=0.32]{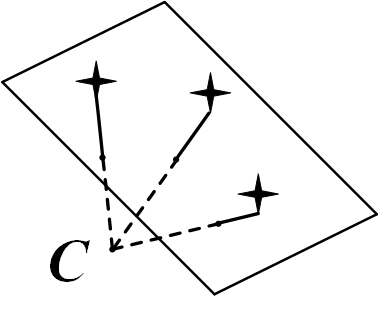}}
	\subfigure[]{
		\label{fig_mapping_to_cube}
		\includegraphics[scale=0.32]{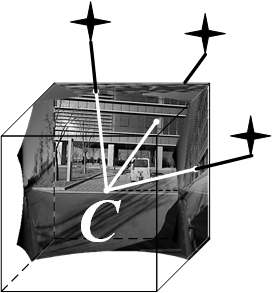}}
	\subfigure[]{
		\label{fig_cubemap_img}
		\includegraphics[scale=0.28]{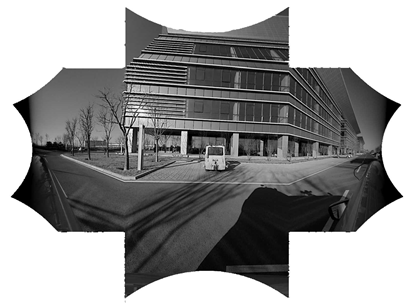}}
	\caption{A demonstration of projecting a fisheye image onto a cube. $C$ are camera centers in the figures. (a) bearing vectors from fisheye image points. (b) project rays onto a single image plane. (c) project rays onto cube. (d) unfolded cubemap image}
	\label{fig_models}
\end{figure}

\subsection{Fisheye Camera Model and the Cubemap Model} 
\label{sub_sec_model}
We choose the omnidirectional camera model from \cite{scaramuzza2006toolbox} to calibrate the fisheye camera in our work. 
By calibrating fisheye camera, we acquire a polynomial which transforms image points into bearing vectors as shown in Fig. \ref{fig_bearing_vectors}. Since the bearing vectors are actually viewing rays, a pinhole image can be acquired by projecting the rays to an image plane with specified camera projection matrix, as in Fig. \ref{fig_mapping_to_plane}. However, projecting on a single pinhole plane would result in a much smaller FoV. To make full use of the large FoV of fisheye camera, we project bearing vectors to multiple image planes. For simplicity, we choose to project onto a cube where each cube face can be seen as an image plane of a virtual pinhole camera with $90^{\circ}$ FoV, and the virtual camera shares the same camera parameters as in Fig. \ref{fig_mapping_to_cube}. After projection, we can get a cubemap image as in Fig. \ref{fig_cubemap_img}.

\subsection{Initialization}
\label{seb_sec_initialization}
Initialization is an important component in SLAM. 
In perspective camera situation, feature matching followed by RANSAC 
is used for solving the fundamental matrix $F$ or the homography matrix $H$ between two frames, and camera motion can be recovered afterwards. 

Although the $F$ and $H$ models are widely used in SLAM for pinhole cameras, it is not possible to calculate them directly on the distorted fisheye images. 
Moreover, the $F$ matrix does not exist because the pinhole camera projection matrix $K$, which is used to derive the $F$ matrix, is undefined for fisheye cameras. 
To make use of the models, we can either rectify the fisheye images into multiple pinhole images that cover the full FoV, and model each pinhole image separately with $F$ or $H$ models, or transform the image points into bearing vectors through the calibrated fisheye model.
For the former approach, we equivalently operate on a multiple pinhole SLAM system as in \cite{forster2017svo}. However, the inter-pinhole correspondence points have to be transformed to the same coordinate first before they can be handled correctly, which increase the complexity.
For the latter approach, the essential matrix model $E$ and $H$ for vectors can be adopted in a vector form, and the intra-pinhole and inter-pinhole correspondences can be handled in a unified framework. In the experiment we find essential matrix model $E$ works for most of the scenarios. Therefore, we represent each measurement as a bearing vector as in \cite{kneip2014opengv}, and apply essential matrix model $E$ for initialization. 
\begin{figure}
	\centering
	\includegraphics[scale=0.35]{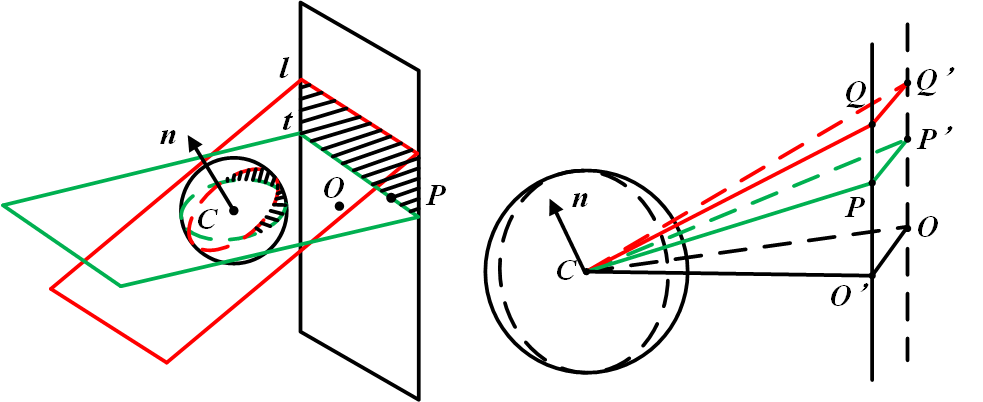} 
	\caption{Threshold of inlier checking in vector-based initialization. The left figure shows the corresponding inlier regions (the shadow areas) between the image plane and the unit sphere. The right figure shows the side view of the left figure. }
	\label{fig_threshold_for_vectors}
\end{figure}

\subsection{Epipolar Constraints on the Unit Sphere}
For SLAM systems, epipolar geometry is used to check whether two points are in correspondence when the $F$ matrix is known, or whether the $F$ assumption is correct when the point correspondences are assumed to be right. To achieve an inlier probability of $95\%$, the following criteria are used for inlier checking,
\begin{equation}
\label{eqn_fundamental}
\dot{\textbf{p}}_2^T F \dot{\textbf{p}}_1 < 3.84\sigma^2
\end{equation}
where $\dot{\textbf{p}}_1$ and $\dot{\textbf{p}}_2$ are homogeneous 
representation of the points $p_1$ and $p_2$ on the images, and $\sigma$ is the variance of the measurement noise (cf. \cite{hartley2003multiple}).
In Eq. \ref{eqn_fundamental}, $F \dot{\textbf{p}}_1$ can be geometrically 
explained as the epipolar line in the second image where $p_2$ belongs. 
Thus a product with $\textbf{p}_2^T$ yields the distance 
of point $p_2$ to the epipolar line. 
Similarly, the essential matrix constraint can be written as
\begin{equation}
\label{eqn_essential}
\textbf{r}_2^T E \textbf{r}_1 = 0
\end{equation}
but $\textbf{r}_1$ and $\textbf{r}_2$ are bearing vectors rather than image points on the plane.
We noticed that since $E$ can be decomposed as $ E = [t]_{\times}R$, where $ E \textbf{r}_1$ indicates
the normal of the epipolar plane,
the formula $\textbf{r}_2^T E \textbf{r}_1$ can be explained as the signed distance 
of $\textbf{r}_2 $ to the epipolar plane. 

As shown in section \ref{sub_sec_model}, for each vector, there is a corresponding image point on the cubemap.
To find inlier threshold for measurements on the unit sphere, we propose to map the well-defined inlier region on image plane to the unit sphere as in Fig. \ref{fig_threshold_for_vectors}.
For simplicity, we only show the process on the front cubemap face. In Fig. \ref{fig_threshold_for_vectors}, the sphere is the unit sphere with point $C$ as camera center, and the plane in black is the front face of cubemap with point $O$ as the center. Line $l$ is the epipolar line from the intersection of the epipolar plane (red) and the image plane (black). $\textbf{n}$ is the normal of epipolar plane.
And we have
\begin{equation}
\label{equn_E_normal}
\textbf{n} = E \textbf{r}_1
\end{equation}
as we mentioned above.
A point $P$ is considered as inlier if the distance to $l$ is within a threshold as Eq. \ref{eqn_fundamental} indicates.
The area within the threshold is represented by the shadow area with line $t$ as boundary. We assume $P$ is on the boundary line $t$ to reveal the boundary conditions. 
For convenience we only draw area under $l$. The area above $l$ can be handled in the same way. The mapped area on unit sphere is also shown in shadow, from which we can see that the corresponding threshold on unit sphere is not uniformly distributed. For area closer to the image plane, the threshold is larger, and for area further from image plane the threshold is smaller. Thus a constant threshold is not reasonable.

To illustrate the geometry relations of the threshold on the image plane and unit sphere, we show the side view of the model in the right figure of Fig. \ref{fig_threshold_for_vectors}. 
In the figure, $OQ'$ is perpendicular to the epipolar line $QQ'$, and parallel to $QO'$ which passes through $P$. 
The plane $OO'QQ'$ corresponds to the image plane in left figure. 
The line segment $CO$, which denotes the focal line, is perpendicular to the image plane and thus perpendicular to $OQ'$. 
Both $P'Q'$ and $PQ$ indicate the threshold on image plane, thus the arc on unit sphere between $QC$ and $CP$ is the inlier region we demand. For simplicity, we notate $\angle PCO'$ as $\phi$, $\angle QCP$ as $\theta$, and length of $QP$ as $th$, which is usually set to 1 pixel. We observe that:
\begin{eqnarray}
\label{eqn_tan_phi_theta}
\tan (\phi + \theta) = \frac{\left\| QO'\right\|}{\left\| CO'\right\|} = \frac{\left\| Q'O\right\|}{\left\| CO'\right\|} = \frac{\left\| th\right\| + \left\| PO'\right\|}{\sqrt{\left\| CO\right\| ^2 + \left\| OO'\right\| ^2}} \\
\label{eqn_tan_phi}
\tan \phi = \frac{\left\| PO'\right\|}{\left\| CO'\right\|} = \frac{\left\| PO'\right\|}{\sqrt{\left\| CO\right\| ^2 + \left\| OO'\right\| ^2}}
\end{eqnarray}
We notice in Eq. \ref{eqn_tan_phi_theta} and \ref{eqn_tan_phi} the length of $OO'$ and $PO'$ are the only unknowns.
And $\vec{OP}$ can be derived immediately since the coordinates of the camera center $O$ and image point $P$ are already known.
To calculate the length of $OO'$ and $PO'$, we can first solve the direction vector $\textbf{e}$ of the epipolar line $QQ'$.
Since $QQ'$ is the intersection of the image plane and epipolar plane, $\textbf{e}$ can be derived by the cross product of normals of the planes.
By making 
\begin{equation}
\vec{z} = \frac{\vec{CO}}{\left\|CO\right\|} = (0, 0, 1)^T
\end{equation}
as the normal of image plane, the direction vector $\textbf{e}$ can be derived by:
\begin{equation}
\vec{e} = \vec{n} \times \vec{z}
\end{equation}
As a result we have:
\begin{eqnarray}
\label{eqn_segment_length}
\left\|OO'\right\| = \frac{\left| \vec{e} \cdot \vec{OP}\right|}{\left\| \vec{e} \right\|} \\
\label{eqn_segment_length_1}
\left\|PO'\right\| = \sqrt{\left\|OP\right\|^2 - \left\|OO'\right\|^2}
\end{eqnarray}
We can derive $\tan (\phi + \theta)$ and $\tan \phi$ by substituting Eq. \ref{eqn_segment_length} and Eq. \ref{eqn_segment_length_1} into Eq. \ref{eqn_tan_phi_theta} and Eq. \ref{eqn_tan_phi}, we have:
\begin{equation}
\label{eqn_tan_theta}
\tan\theta = \frac{\tan (\phi + \theta) - \tan\phi}{1+\tan (\phi + \theta) \tan\phi}, 
\label{eqn_cos_theta}
\sin\theta = \frac{\tan\theta}{\sqrt{\tan^2\theta+1}}
\end{equation}
Then from Eq.\ref{equn_E_normal} and Eq.\ref{eqn_essential}, we get our inlier metric for the unit shpere as:
\begin{equation}
\left | \frac{\textbf{r}_2^T E \textbf{r}_1}{\left\|\textbf{r}_2\right\| \left\|E \textbf{r}_1\right\|} \right | = \left | \frac{\textbf{r}_2^T \textbf{n}}{\left\|\textbf{r}_2\right\| \left\|\textbf{n}\right\|} \right | \leq \left |\cos (\frac{\pi}{2}\pm\theta) \right | = \left | \sin\theta \right |
\end{equation}

\subsection{Optimization}
\label{seb_sec_optimization}
To perform optimizations in vector-based vision systems, several metrics have been proposed.
\cite{kneip2014opengv} proposes to minimize the angular error between bearing vectors, and \cite{pagani2011structure} studies different metrics and shows that the tangential error has the best performance.
Zhang et al. \cite{zhang2016benefit} evaluate the above metrics as well as a vector difference with a semi-direct VO\cite{forster2017svo}.
Inspired by the multi-pinhole nature of cubemap, we propose to minimize reprojection errors of all cube faces as a multi-camera system. 

The multi-camera model is used extensively in previous multiple-camera SLAM systems 
\cite{forster2017svo,urban2016multicol,furgale2013toward,lee2013motion}. 
In the multi-camera model, a body frame $B$ rigidly attached to camera frames is set as 
the reference frame. 
Transformations $T_{C_i B}$ from body frame to camera local frames $C_i$ can be 
obtained by extrinsic calibration, where $i$ represent the camera index. 
In cubemap model, different faces are equivalent to pinhole cameras as in section \ref{sub_sec_model}. 
We set the front-facing virtual camera as the body frame, 
and all the pinhole cameras are transformed to the body frame 
by a rotation $R_{C_i B}$. The projection model of cubemap is:
\begin{equation}
\label{eqn_projection}
u = K R_{C_i B} T_{BW} P
\end{equation}
where $P=(x,y,z)^T$ is the 3D point in world frame, 
$T_{BW}$ is transformation from world frame to body frame, and $u$ is the local point coordinate in the image coordinate of each cubemap face.  

We can represent $T\in SE_3$ with $\xi = (\phi^T, \rho^T)^T$ \cite{barfoot2017state} and expand Eq. $\ref{eqn_projection}$ into:
\begin{equation}
u=K P_1, P_1 = R_{C_i B} P_2, P_2 = T_{BW} P
\end{equation}
The Jacobian of the measurement $u$ to camera pose $T$ therefore can be 
derived according to the chain rule:
\begin{equation}
\label{J_xi}
J_{\xi} = -\frac{\partial u}{\partial P_1} \cdot R_{C_i B} \cdot \left[-P_2^{\wedge}, I_{3\times 3}\right]
\end{equation}
where $P_2^{\wedge}$ is the skew-symmetric matrix of $P_2$. The Jacobian for map point position is given by:
\begin{equation}
J_p = -\frac{\partial u}{\partial P_1} \cdot \frac{\partial P_1}{\partial P}
= -\frac{\partial u}{\partial P_1} R_{C_i B} R_{BW}
\end{equation}
where $R_{BW}$ is the rotation part of $T_{BW}$.

To find the best metric for CubemapSLAM, we thoroughly evaluated the metrics.
For convenience we keep the notation used in \cite{zhang2016benefit}, where the angular metrics are denoted as $r_{a1}$ and $r_{a2}$, and the tangential metric and vector difference metric are denoted as $r_t$ and $r_f$ respectively. The multi-camera model based metric is denoted as $r_u$. 
We compute the ATE RMSE(Absolute trajectory error) \cite{sturm2012benchmark} of the system with different metrics in pose optimization. The evaluation is performed on a long straight track with local bundle adjustment disabled.
In the result, $r_t$ and $r_f$ achieve errors as $11.27m$ and $20.53m$ and fail to keep the scale of the map, and $r_{a1}$ and $r_{a2}$ fail quickly after initialization. In contrast $r_u$ achieves the most accurate result as $1.03m$, which indicates that the multi-pinhole model is more suitable for our system.

\section{Experiments}
\label{sec_expriment}
We evaluate the performance of our system in the multi-fisheye dataset Lafida\cite{urban2017lafida} as well as a dataset collected in large outdoor environments with our autonomous vehicle. 
In the Lafida dataset, we evaluate CubemapSLAM and 
Multicol-SLAM \cite{urban2016multicol} on both accuracy and robustness. 
Then on the dataset collected by ourselves, 
we first evaluate the systems under two types of camera settings as in section \ref{sec_baseline_comp}. We further investigate the effect of different mounting positions of the camera and loop closure. The result shows that our CubemapSLAM system performs consistently more robustly in all the experiments than the other ones, and it provides competitive accuracy.

\subsection{Dataset}
\label{sec_dataset}
The Lafida dataset\cite{urban2017lafida} is a multi-fisheye camera dataset collected for evaluating multi-fisheye SLAM systems. There are 6 sequences in total, which are $in\_dynamic$, $in\_static$, $out\_static$, $out\_static2$, $out\_rotation$ and $out\_large\_loop$ captured from three rigidly mounted fisheye cameras. All the cameras share the same resolution of $754\times 480$ pixels. 
As for our dataset, we equip the vehicle with two types of cameras: a pinhole camera with $80^{\circ}$ FoV, and a fisheye camera with $190^{\circ}$ FoV. Note that the pinhole camera and fisheye camera share the same model of sensor chip but use different lenses, so the pinhole and raw fisheye image have the same resolution of $1280\times 720$ pixels. 

The experiments on our dataset contain two camera settings. In the first setting, a pinhole camera and a fisheye are mounted at the frontal part of the vehicle (both facing front), and the other fisheye camera is mounted at the left side of the vehicle (facing left). In this setting, we choose various 
routes, including a large loop around an industrial park 
($loop1$ sequence), 
a smaller loop inside the park with sharp turns ($loop2$ sequence), 
a large u-turn on the $loop1$ sequence ($uturn$ sequence), 
a large sequence in a town with no loop but with traffic lights and traffic jams ($town$ sequence), and a loopy route in an outdoor parking lot ($parkinglot$ sequence). 
To further investigate the performance of the lateral mounting cameras, we create a second setting where a fisheye camera and a pinhole camera are both mounted laterally. We travel along the routes of $loop1$, $town1$ and $parkinglot$ and recollect the data under the new camera setting. We name the collected data $loop1\_c\_clockwise$, $loop1\_clockwise$, $town\_1$ and $parkinglot\_1$. $loop1\_c\_clockwise$ and $loop1\_clockwise$ share the same route but drive in opposite directions.
For all the sequences, D-GPS is used as groundtruth.

\subsection{Baseline Comparison}
\label{sec_baseline_comp}
We first compare our system with Multicol-SLAM \cite{urban2016multicol} on the Lafida dataset \cite{urban2017lafida}, where Multicol-SLAM is sufficiently tested and well performed. For fairness, Multicol-SLAM is configured with one fisheye camera which is the same one as CubemapSLAM. 
For a comprehensive comparison, we set the resolution of the faces as $450\times 450$, $550\times 550$ and $650\times 650$ pixels, respectively. In the experiment, both of the systems are configured to extract $2000$ features, which we consider it is enough and representative for the dataset considering the image resolution.

On our dataset, the CubemapSLAM operates on the front (\textbf{Cube-F}) and left(\textbf{Cube-L}) fisheye cameras. 
We simply set the face resolution as $650\times 650$ pixels.
The first baseline comparision is to perform ORB-SLAM\cite{mur2017orb} on rectified fisheye images from front (\textbf{ORB-Rect-F}) and left (\textbf{ORB-Rect-L}) cameras. The rectified images are set to $100^{\circ}$ FoV with a resolution of $775\times 775$ pixels which share the same focal length with cubemap virtual cameras.
Another baseline approach is to perform ORB-SLAM\cite{mur2017orb} on pinhole images from front(\textbf{ORB-Pin-F}) and left cameras(\textbf{ORB-Pin-L}).
We also tested Multicol-SLAM \cite{urban2016multicol} on the collected dataset. 
However, we find Multicol-SLAM fails soon after initialization stage in most of the sequences as it does in Lafida $out\_large\_scale$ sequence, so we do not include trajectories of Multicol-SLAM in result comparison.

We first test the systems with loop closing thread disabled to evaluate system performance with only VO. 
In the experiments, all the systems are configured to extract $3000$ features per image.
For each output trajectory, we align it with the ground truth by a 7-DoF transformation since the scale is unknown. After that, we compute the ATE RMSE \cite{sturm2012benchmark} of each trajectory for comparison. We also evaluate the tracking and mapping quality of all the systems by measuring the average number of tracked keypoints in each sequence. 
Note that for all the entries in the tables, we add a mark of $lost$ in the entry if the system gets lost after finishing more than half of the sequence, and we add a mark of X if the system gets lost soon after initialization. 
\begin{figure*}
	\centering
	\subfigure[in\_dynamic]{
		\label{fig_traj_compare_in_dynamic}
		\includegraphics[width=0.31\linewidth]{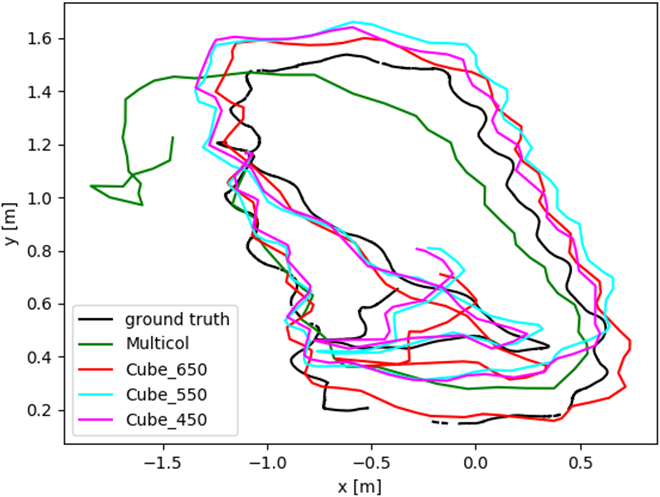}}
	\subfigure[in\_static]{
		\label{fig_traj_compare_in_static}
		\includegraphics[width=0.31\linewidth]{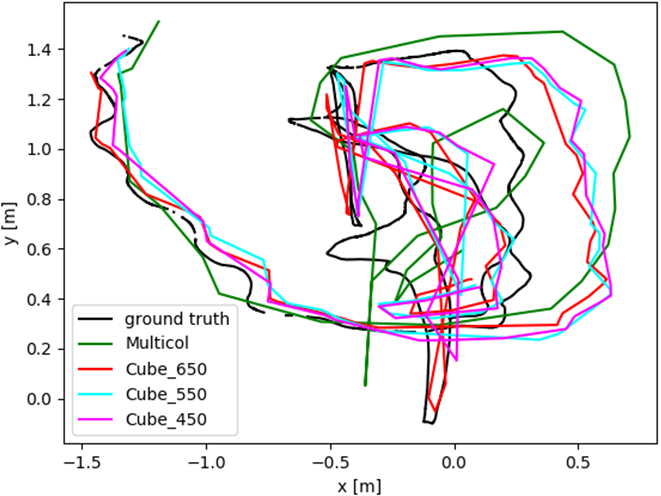}}
	\subfigure[out\_static]{
		\label{fig_traj_compare_out_static}
		\includegraphics[width=0.31\linewidth]{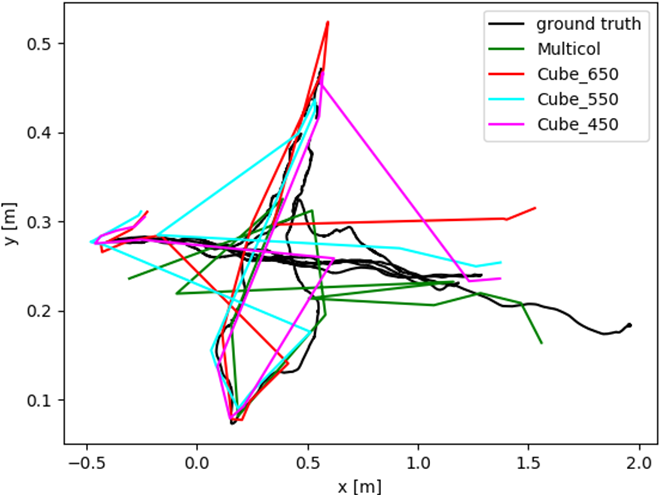}}
	\subfigure[out\_static2]{
		\label{fig_traj_compare_out_static2}
		\includegraphics[width=0.31\linewidth]{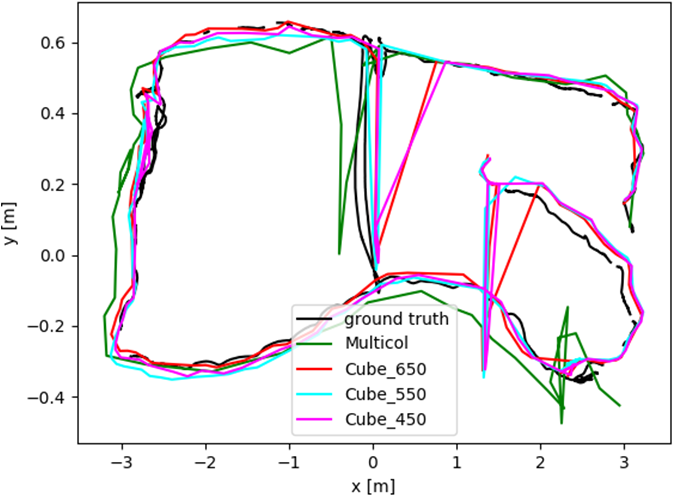}}
	\subfigure[out\_rotation]{
		\label{fig_traj_compare_out_rotation}
		\includegraphics[width=0.31\linewidth]{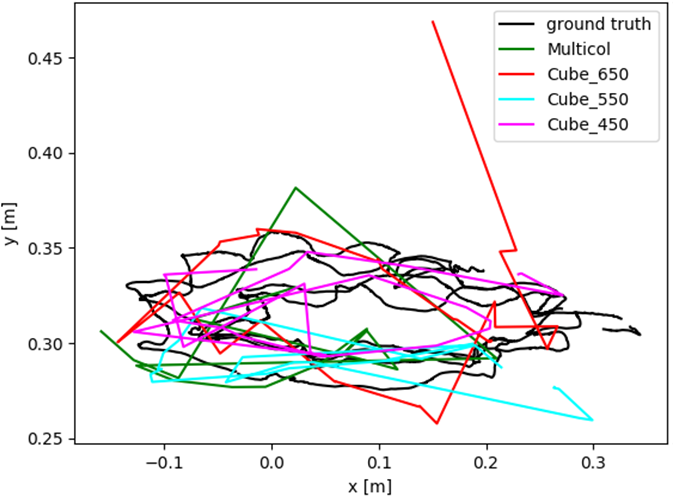}}
	\subfigure[out\_large\_loop]{
		\label{fig_traj_compare_out_large_loop}
		\includegraphics[width=0.31\linewidth]{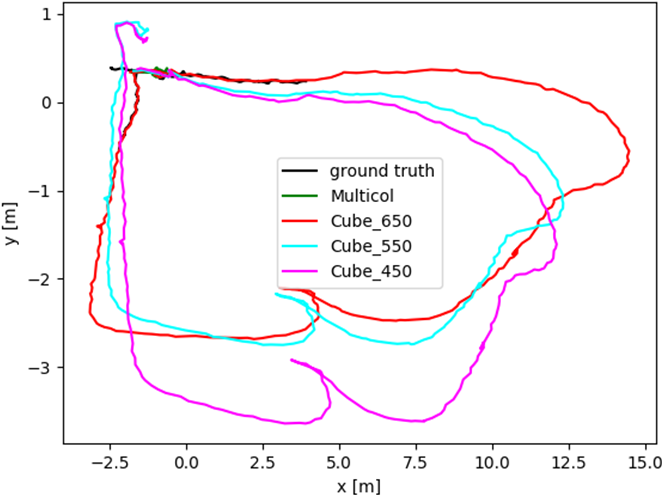}}
	\caption{Trajectories on Lafida dataset\cite{urban2017lafida} aligned to groundtruth with a 7-DoF transformation of Multicol-SLAM and CubemapSLAM with resolution of face as $450\times 450$, $550\times 550$ and $650\times 650$ pixels, respectively.}
	\label{fig_traj_compare_lafida}
\end{figure*}

\begin{table}
	\caption{ATE RMSE and tracked frames(over all frames) on Lafida dataset\cite{urban2017lafida} (m)}
	\label{table_lafida_ate_rmse}
	\centering
	\begin{tabular}{|l|l|l|l|l|l|l|}
		\hline
		& in\_dynamic & in\_static & out\_static & out\_static2 & out\_rotation & out\_large\_loop \\
		\hline
		Multicol & \makecell{0.78\\(880/899)} & \makecell{0.32\\(1001/1015)} & \makecell{0.07\\(\textbf{726}/755)} & \makecell{0.31\\(1314/1642)} & \makecell{0.05\\(397/779)} & \makecell{0.06\\(202/3175)}  \\
		\hline
		\makecell{Cubemap\\$650\times 650$} & \makecell{\textbf{0.17}\\(\textbf{893}/899)} & \makecell{\textbf{0.15}\\(997/1015)} & \makecell{\textbf{0.02}\\(722/755)} & \makecell{0.14\\(1604/1642)} & \makecell{0.06\\(253/779)} & \makecell{\textbf{0.16}\\(3111/3175)} \\
		\hline
		\makecell{Cubemap\\$550\times 550$} & \makecell{0.28\\(\textbf{893}/899)} & \makecell{0.16\\(\textbf{1006}/1015)} & \makecell{0.03\\(717/755)} & \makecell{0.15\\(1604/1642)} & \makecell{0.10\\(399/779)} & \makecell{0.44\\(\textbf{3132}/3175)} \\
		\hline
		\makecell{Cubemap\\$450\times 450$} & \makecell{0.24\\(\textbf{893}/899)} & \makecell{0.17\\(\textbf{1006}/1015)} & \makecell{0.03\\(716/755)} & \makecell{\textbf{0.13}\\(\textbf{1605}/1642)} & \makecell{\textbf{0.04}\\(\textbf{743}/779)} & \makecell{0.39\\(3129/3175)} \\
		\hline
	\end{tabular}
\end{table}

\subsection{Results on Lafida Dataset}
We carefully evaluate both systems on all the six sequences, and the qualitative and quantitative results are shown in Table \ref{table_lafida_ate_rmse} and Fig.\ref{fig_traj_compare_lafida}. The results show that the CubemapSLAM performs better than  Multicol-SLAM in most sequences and the performance is stable with various face size. Also it should be noted that in $out\_large\_scale$ although the error of Multicol-SLAM is slightly lower, the number of tracked frames are significantly less than ours. We've tested Multicol with several different start points for fairness, but the results do not show much difference. We notice Multicol-SLAM usually fails when the camera motion becomes large. However, large motions are very common in large-scale outdoor dataset.
\begin{figure*}
	\centering
	\subfigure[loop1]{
		\label{fig_traj_compare_fangshan2_l1}
		\includegraphics[width=0.31\linewidth]{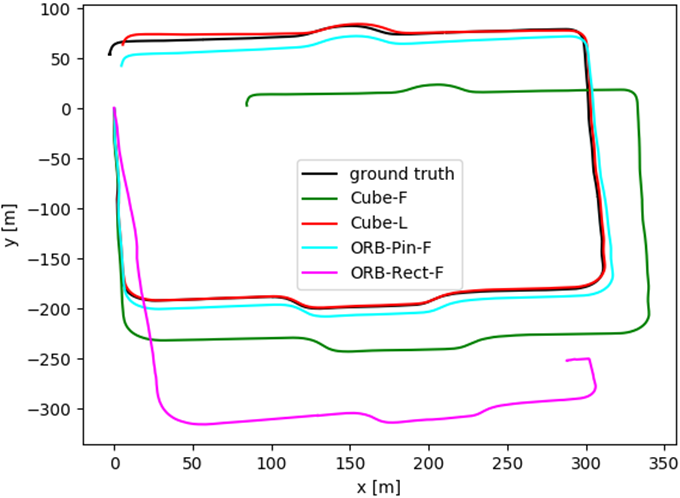}}
	\subfigure[loop2]{
		\label{fig_traj_compare_fangshan3_l5}
		\includegraphics[width=0.31\linewidth]{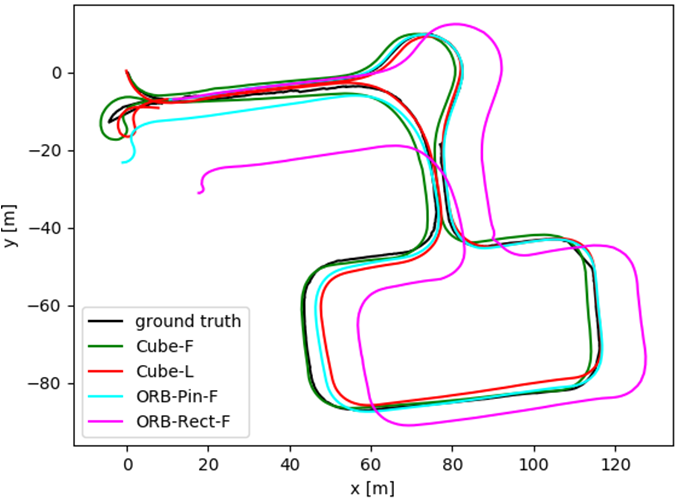}}
	\subfigure[uturn]{
		\label{fig_traj_compare_fangshan3_l1_uturn}
		\includegraphics[width=0.31\linewidth]{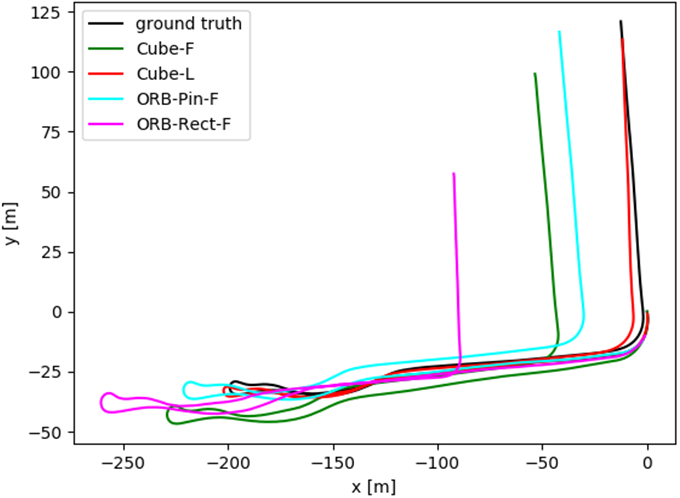}}
	\subfigure[town]{
		\label{fig_traj_compare_fangshan3_town}
		\includegraphics[width=0.31\linewidth]{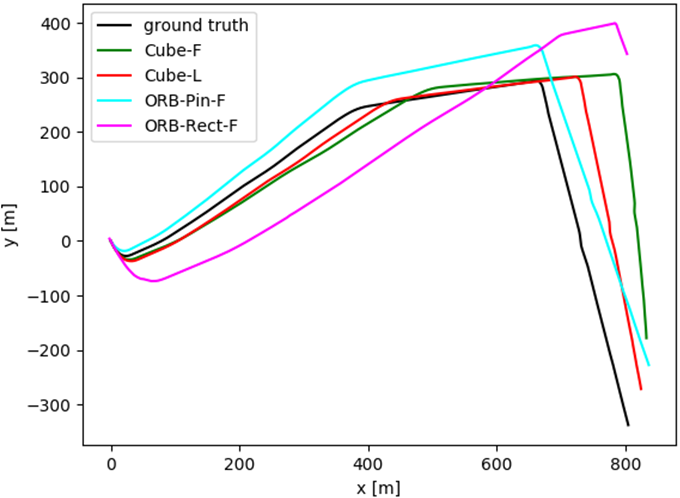}}
	\subfigure[parkinglot]{
		\label{fig_traj_compare_fangshan3_parking_lot}
		\includegraphics[width=0.31\linewidth]{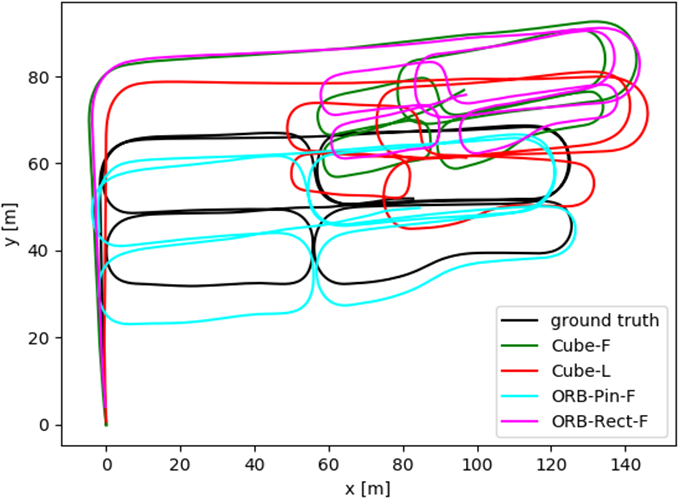}}
	\caption{Trajectories aligned to groundtruth with a 7-DoF transformation of the \textbf{Cube-F}, \textbf{Cube-L}, \textbf{ORB-Pin-F} and \textbf{ORB-Rect-F}.}
	\label{fig_traj_compare}
\end{figure*}

\begin{table}
	\caption{ATE RMSE (m) and Average Number of Tracked Points (pt)}
	\label{table_ate_rmse}
	\centering
	\begin{tabular}{|l|l|l|l|l|l|l|l|l|}
	    \hline
	    & \multicolumn{4}{|c|}{ATE RMSE} & \multicolumn{4}{|c|}{Average Number of Tracked Points} \\
		\hline
		& Cube-F & Cube-L & \makecell{ORB-\\Pin-F} & \makecell{ORB-\\Rect-F} & Cube-F & Cube-L & \makecell{ORB-\\Pin-F} & \makecell{ORB-\\Rect-F} \\
		\hline
		loop1 	& 22.73 & \textbf{3.12} & 3.92 & 121.14 & 194.21 & 198.73 & \textbf{206.84} & 190.73 \\
		\hline
		loop2 & 2.68 & \textbf{2.26} & 2.54(lost) & 6.66(lost) & 210.64 & 229.35 & \textbf{232.28} & 201.45  \\
		\hline
		uturn & 12.14 & \textbf{2.22} & 7.99 & 27.85 & 196.78 & \textbf{236.65} & 176.74 & 150.93 \\
		\hline
		town & 57.32 & 23.58 & \textbf{13.41} & 207.15 & 256.06 & \textbf{338.06} & 222.50 & 207.29 \\
		\hline
		parkinglot & 16.47 & 9.54 & \textbf{2.29} & 16.92 & 170.34 & \textbf{182.45} & 148.36 & 164.34 \\
		\hline
	\end{tabular}
\end{table}

\subsection{Results of Setting1}
The ATE RSME results for each method can be found in Table \ref{table_ate_rmse}, and the comparison of the trajectories from different systems are shown in Fig. \ref{fig_traj_compare}. 
In all the sequences, the ATE error of \textbf{ORB-Rect-F} is significantly larger than the other methods, which is the consequence of both a reduced FoV from the full fisheye image and a lower angular resolution than that of the pinhole camera. \textbf{ORB-Pin-F} has a low ATE RMSE in most of the sequences due to its higher angular resolution than fisheye image. However, in the $loop2$ sequence, \textbf{ORB-Pin-F} fails to complete the entire trajectory due to a drastic turn at the end of the sequence. In contrast, both \textbf{Cube-F} and \textbf{Cube-L} successfully complete all the sequences including the difficult $loop2$ with the help of a larger FoV. In addition, we find that in $loop1$, $loop2$ and $uturn$ sequences, \textbf{Cube-L} achieves the best result. In the $town$ and the $parkinglot$ sequences, as the feature points are relatively far from the camera, \textbf{ORB-Pin-F} outperforms \textbf{Cube-L} by a small margin, but we will show that the gap is significantly reduced after loop closure.

For the number of tracked keypoints, as in Table \ref{table_ate_rmse}, \textbf{Cube-L} tracks the most points in $uturn$, $town$ and $parkinglot$, and performs close to \textbf{ORB-Pin-F} in $loop1$ and $loop2$. 
Note that besides the advantage of better tracking quality, more tracked keypoints also contributes to a denser and more structural map.
\begin{table*}
	\caption{ATE RMSE (m) and Average Number of Tracked Points (pt)}
	\label{table_fangshan6_ate_rmse}
	\centering
	\begin{tabular}{|l|l|l|l|l|l|l|l|}
	    \hline
	    & \multicolumn{3}{|c|}{ATE RMSE} & \multicolumn{3}{|c|}{Average Number of Tracked Points} \\
		\hline
		& Cube-L & ORB-Pin-L & ORB-Rect-L & Cube-L & ORB-Pin-L & ORB-Rect-L  \\
		\hline
		loop1\_c\_clockwise & \textbf{9.84} & 15.60 &  X & \textbf{280.01} & 215.06 &  241.97 \\
		\hline
		loop1\_clockwise & \textbf{8.94} & X & X & \textbf{136.99} & X & X \\
		\hline
		town\_1 & \textbf{14.75} & 16.20(lost) & 6.94(lost) & 366.14 & \textbf{379.21} & 375.09 \\
		\hline
		parkinglot\_1 &\textbf{5.44} &21.74(lost) & X & \textbf{193.80} & 191.01 & 152.76\\
		\hline
	\end{tabular}
\end{table*}

\begin{figure*}
	\centering
	\subfigure[loop1\_1\_c\_clockwise]{
		\label{fig_traj_compare_fangshan6_l1_inner}
		\includegraphics[scale=0.3]{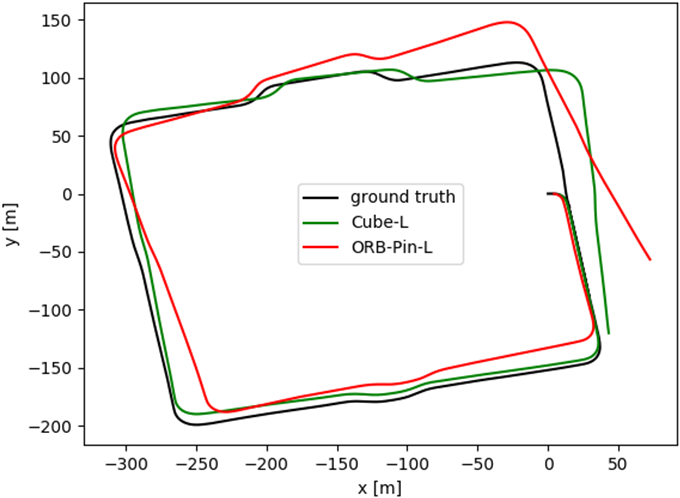}}
	\subfigure[loop1\_1\_clockwise]{
		\label{fig_traj_compare_fangshan6_l1_outer}
		\includegraphics[scale=0.3]{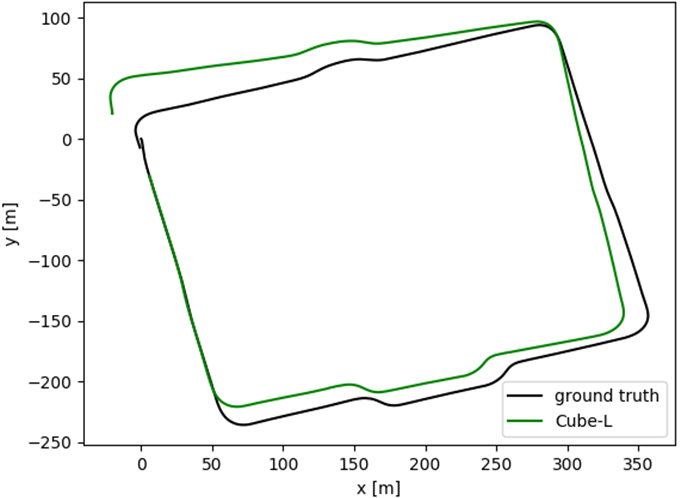}}
	\subfigure[town\_1]{
		\label{fig_traj_compare_fangshan6_doudian_town}
		\includegraphics[scale=0.3]{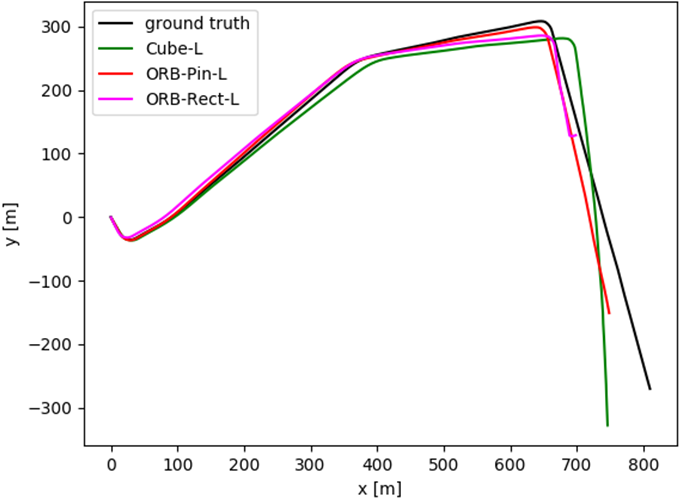}}
	\subfigure[parkinglot\_1]{
		\label{fig_traj_compare_fangshan6_parking_lot}
		\includegraphics[scale=0.3]{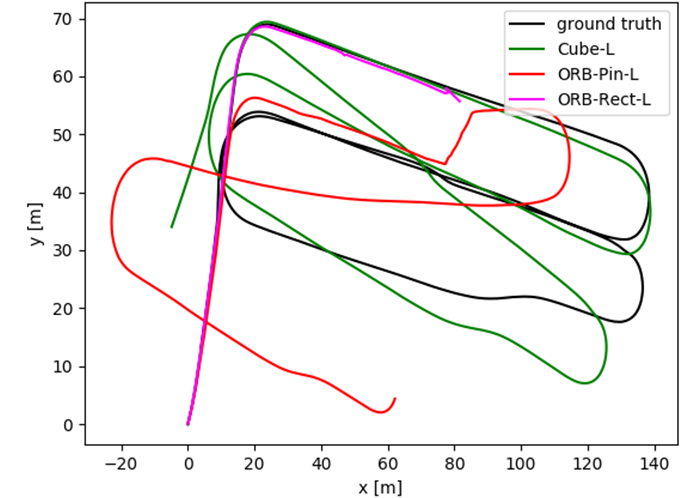}}
	\caption{Trajectories aligned to groundtruth with a 7-DoF transformation of the \textbf{Cube-L}, \textbf{ORB-Pin-L} and \textbf{ORB-Rect-L}.
	}
	\label{fig_traj_compare_left}
\end{figure*}

\begin{figure*}
	\centering
	\subfigure[parkinglot]{
		\label{fig_traj_compare_fangshan3_parking_lot_loop_closing}
		\includegraphics[scale=0.3]{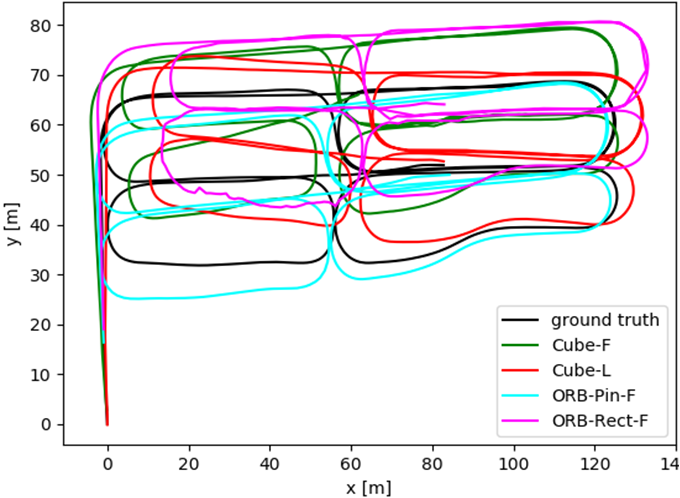}}
	\subfigure[parkinglot\_1]{
		\label{fig_traj_compare_fangshan6_parking_lot_loop_closer}
		\includegraphics[scale=0.3]{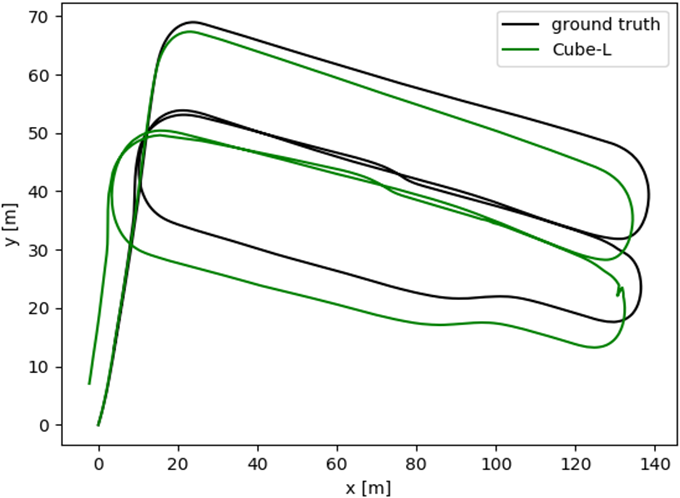}}
	\caption{Trajectories aligned to groundtruth with a 7-DoF transformation of the Cubemap and ORB-SLAM with loop closing.
	}
	\label{fig_traj_compare_left_loop_closing}
\end{figure*}

\begin{figure*}
	\centering
	\label{fig_traj_compare_faster_frame_rate_cube}
	\includegraphics[scale=0.45]{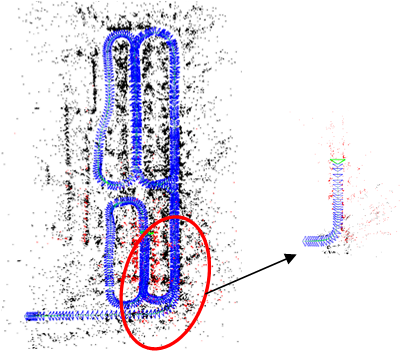}
	\caption{Qualitative results of \textbf{Cube-F} (left) and \textbf{ORB-Pin-F} (right) on the $parkinglot$ sequence with a lower frame rate by choosing one image from every three images. 
	}
	\label{fig_traj_compare_faster_frame_rate}
\end{figure*}

\subsection{Results of Setting2}
To make the comparison fair for the lateral mounting cameras, we mount both types of cameras towards left and evaluate the performance respectively. Quantitative and qualitative results are shown in Table \ref{table_fangshan6_ate_rmse} and Fig. \ref{fig_traj_compare_left}. In all the sequences, \textbf{Cube-L} outperforms the other two methods by a large margin. In the clockwise sequence where the cameras look outwards the park, both \textbf{ORB-Pin-L} and \textbf{ORB-Rect-L} get lost soon after initialization due to lack of texture and occlusions by objects close-by. We therefore do not compute the error and replace each field with a X. Also in $town\_1$, both \textbf{ORB-Rect-L} and \textbf{ORB-Pin-L} get lost before finishing the sequence due to occlusion from cars passing by.
In addition, we list the number of average tracked points in Table 
\ref{table_fangshan6_ate_rmse}, in which \textbf{Cube-L} achieves better overall performance than the other systems.

\begin{table*}
	\caption{ATE RMSE with Loop Closure (m) on parkinglot and parkinglot\_1 Sequences}
	\label{table_fangshan3_ate_rmse_with_loop_closer}
	\centering
	\begin{tabular}{|l|l|l|l|l|l|l|l|}
	    \hline
	    & \multicolumn{4}{|c|}{parkinglot} & \multicolumn{3}{|c|}{parkinglot\_1} \\
	    \hline
		& Cube-F & Cube-L & \makecell{ORB-\\Pin-F} & \makecell{ORB-\\Rect-F} & Cube-L & \makecell{ORB-\\Pin-L} & \makecell{ORB-\\Rect-L} \\
		\hline
		w/o loop closing & 16.47 & 9.54 & \textbf{2.29} & 16.92 & \textbf{5.44} &21.74(lost) & X \\
		\hline
		w/ loop closing & 4.47 & 3.02 & \textbf{1.69} & 4.73 & \textbf{2.41} & X & X \\
		\hline
	\end{tabular}
\end{table*}

\subsection{Results of Loop Closing}
To evaluate system performance with loop closing thread enabled, we test the performances of the systems in both $parkinglot$ and $parkinglot\_1$ sequences with and without loop closing.
Results show that errors of \textbf{Cube-F} and \textbf{Cube-L} are greatly reduced and getting comparable to \textbf{ORB-Pin-F} with loop closing (see Table 4), and the robustness consistently outperform the rest. A qualitative result is shown in Fig. \ref{fig_traj_compare_left_loop_closing}.
To further reveal the advantage of a large FoV camera, we test \textbf{ORB-Pin-F} and \textbf{Cube-F} in the $parkinglot$ sequence with a reduced frame rate by choosing one image out of three in the sequences. We find that \textbf{ORB-Pin-F} is hard to initialize and fails soon after initialization, while \textbf{Cube-F} is able to initialize fast, track stably, and successfully perform loop closing. A qualitative result of the trajectories is shown in Fig. \ref{fig_traj_compare_faster_frame_rate}.

\section{Conclusions}
This work presents a novel CubemapSLAM system that incorporates the cubemap model into the state-of-the-art feature based SLAM system. The cubemap model utilizes the large FoV of fisheye camera without affecting the performance of feature descriptors. In addition, CubemapSLAM is efficiently implemented and can run in real time. In the experiments, we extensively evaluate our systems in various challenging real-world cases and prove that our CubemapSLAM solution is consistently more robust than other approaches without losing accuracy. We also discover that by optimizing the mounting position of the fisheye camera and enabling the loop closing thread, CubemapSLAM can achieve even better accuracy than pinhole cameras, despite the limited angular resolution of the sensor. Overall, we provide an efficient and practical fisheye SLAM solution. Future work includes extending the cubemap model to stereo or multiple camera setting to further improve robustness as well as recover the absolute scale. 
The source code is available at \url{https://github.com/nkwangyh/CubemapSLAM}.

%
%
%


\end{document}